\pdfoutput=1
\documentclass[11pt]{article}

\usepackage[final]{acl}

\usepackage{times}
\usepackage{latexsym}

\usepackage[T1]{fontenc}
\usepackage[utf8]{inputenc}

\usepackage{microtype}

\usepackage{inconsolata}

\definecolor{Gray}{gray}{0.9}
\definecolor{LightCyan}{rgb}{0.88,1,1}

\usepackage{graphicx}
\usepackage{tikz}
\usepackage{xspace}
\usepackage{xcolor,pifont}
\usepackage{url}            % simple URL typesetting
\usepackage{booktabs}       % professional-quality tables
\usepackage{amsfonts}       % blackboard math symbols
\usepackage{nicefrac}       % compact symbols for 1/2, etc.
\usepackage{comment}

\usepackage{amsmath}
\usepackage{amssymb}
\usepackage{tablefootnote}
\usepackage{enumitem}
\usepackage{caption}
\usepackage{subcaption}

\usepackage{multirow}
\usepackage{makecell} 
\usepackage{ragged2e}
\usepackage{comment}

\usepackage{pifont}

\usepackage{textgreek}

\usepackage{tipa}
\usepackage{lscape}
\usepackage{tikz,lipsum}
\usepackage[most]{tcolorbox}

\usepackage{hyperref}
\usepackage{times}
\usepackage{latexsym}
\usepackage{xspace,mfirstuc,tabulary}
\usepackage{adjustbox}

\usepackage{color, colortbl}
\usepackage{xcolor}
\usepackage{enumitem}
\usepackage{amssymb} % for checkmark
\usepackage{pifont} % for xmark
\usepackage{multirow}

\newcommand{\insertafrimgsmtable}{
    \begin{table*}[h!]
        \small\centering
        \resizebox{\textwidth}{!}{%
          \begin{tabular}{r|lccc|ccccc|cc|cc}
          \toprule
          & \textbf{Base} & \textbf{Prompt} & \textbf{Data} & \textbf{\# Synth}  & \textbf{Yor} & \textbf{Ibo} & \textbf{Hau} & \textbf{Swa} & \textbf{Zul} & \textbf{Eng} & \textbf{Fra} & \textbf{Avg} & \textbf{Avg w/o} \\
          & \textbf{Model} & \textbf{Masking} & & \textbf{Samples} & & & & & & & & & \textbf{eng \& fra} \\
          \midrule
          1 & GPT-4 (gpt-4-0125-preview) & - & - & -  & 26.0 & 18.0 & 34.0 & 50.4 & 26.0 & 73.6 & 54.8 & 40.4 & 30.9 \\
          2 & Llama 3.1 8b Base & - & - & - &  5.2 & 3.2 & 5.6 & 7.2 & 3.6 & 17.6  & 13.6 & 5.0 & 8.0 \\
          3 & Llama 3.1 8b Instruct & - & - & - & 4.8 & 8.0 & 11.2 & 42.0 & 7.2 & 38.8  & 45.2 & 22.5 & 14.6 \\
          4 & Llama 3.1 70b Instruct & - & - & - & 20.0 & 37.2 & 48.0 & 76.4 & 24.4 & 74.4 & 76.4 &  \\
          \midrule
          &&&&& \multicolumn{7}{c|}{Monolingual SFT} \\
          5 & Llama 3.1 8b Instruct & \xmark & Open Instruct Translated & - & 27.6 & 21.6 & 29.2 & 45.2 & 25.2 & - & - & - & 29.8 \\
          \midrule
          &&&&& \multicolumn{7}{c|}{Multilingual SFT} \\
          6 & Llama 3.1 8b Instruct & \xmark  & \Afripersonainst  & 10,000  & 19.2 & 17.6 & 28.0 & 40.8 & 15.6 & 67.6  & 55.6 & 34.9 & 24.2 \\
          7 & Llama 3.1 8b Instruct & \xmark & \Afripersonainst &  20,000  & 19.2 & 22.0 & 32.0 & 43.6 & 20.8 & 59.6  & 45.6 & 34.7 & 27.5 \\
          8 & Llama 3.1 8b Instruct & \xmark & \Afripersonainst & 30,000  & 25.6 & 22.0 & 32.4 & 54.4 & 24.0 & 65.6  & 51.6 & 39.4 & 31.8 \\
          9 & Llama 3.1 8b Instruct & \cmark & \Afripersonainst & 30,000 & 19.6 & 23.6 & 32.8 & 50.8 & 21.6 & 69.2  & 53.2 & 38.7 & 29.7 \\
          10 & Llama 3.1 8b Instruct & \xmark  & BigMath Translated &  30,000  & 10.4 & 9.6 & 14.0 & 13.6 & 7.2 & 25.2  & 18.4 & 11.2 & 8.96 \\
          11 & Llama 3.1 8b Instruct & \xmark  & Open Instruct Translated &  30,000  & 46.4 & 37.6 & 49.6 & 64.0 & 36.0 & 74.0  & 52.0 & 51.4 & 46.8 \\
          12 & Llama 3.1 8b Instruct & \xmark & All Data &  60,000  & 46.8 & 42.4 & 59.2 & 69.2 & 44.0 & 78.0  & 64.0 & 57.7 & 52.3 \\
          \midrule
          13 & Llama 3.1 8b Base & \xmark & \Afripersonainst & 30,000  & 20.0 & 21.6 & 27.2 & 37.6 & 15.2 & 52.8  & 38.8 & 30.5 & 24.3 \\
          14 & Llama 3.1 8b Base & \xmark & Open Instruct Translated & 30,000  & 42.8 & 35.6 & 50.0 & 56.4 & 38.0 & 68.4 & 51.6 & 48.9 & 44.6   \\
          15 & Lugha Llama 3.1 8b Base & \xmark & Open Instruct Translated & 30,000  & 31.2 & 30.4 & 45.6 & 41.6 & 31.2 & 56.0 & 35.6 & 38.8 & 36.0 \\
          \midrule
      \bottomrule
      \end{tabular}
    }
      \caption{\textbf{AfriMGSM Accuracy:} Performance of different model, data configurations on the AfriMGSM benchmark. For the monolingual SFT section, we train individual models for each language and evaluate that model on that language only while the multilingual models were trained data from all languages.}
      \label{tab:afrigmsmresult}
    \end{table*}
}

\newcommand{\insertafrimgsmtablezeroshot}{
    \begin{table*}[h!]
        \small\centering
        \resizebox{\textwidth}{!}{%
          \begin{tabular}{ll|rrrrrrrrrr|r}
          \toprule
          \textbf{Base} & \textbf{\# Synth}  & \textbf{Ewe} & \textbf{Kin} & \textbf{Lin} & \textbf{Lug} & \textbf{Orm} & \textbf{Sna} & \textbf{Sot} & \textbf{Twi} & \textbf{Xho} & \textbf{Wol} & \textbf{Avg}  \\
          \textbf{Model} & \textbf{Samples} & & & & & & & & & \\
          \midrule
          Llama 3.1 8b Base & - & 1.2 & 4.0 & 0.8 & 5.2 & 2.0 & 3.2 & 2.4 & 3.2  & 2.0 & 0.8 & 2.5  \\
          Llama 3.1 8b Instruct &  - & 0.8 & 6.4 & 4.4 & 8.0 & 4.8 & 6.0 & 5.2 & 3.6  & 4.8 & \textbf{5.2} & 4.9 \\
          Llama 3.1 8b Instruct -- \Afripersonainst &  30,000  & 3.6 & 5.6 & 6.0 & 6.0 & 2.4 & 7.2 & 3.2 & 2.0  & 14.0 & 1.6 & 5.2 \\
          Llama 3.1 8b -- Open Instruct Translated &  30,000 & 4.4 & 6.0 & 5.2 & 8.0 & 7.6 & \textbf{10.4} & 5.6 & 4.0 & 28.0  & 3.6 & 8.3 \\
          Llama 3.1 8b -- All Data &  60,000 & \textbf{6.0} & \textbf{8.4} & \textbf{8.4} & \textbf{9.6} & 5.6 & 8.8 & \textbf{7.2} & \textbf{4.4} & \textbf{30.8}  & 2.4 & 9.2 \\
          \midrule
      \bottomrule
      \end{tabular}
    }
      \caption{\textbf{AfriMGSM -- Zero Shot Results:} Performance of different model configurations on unseen languages during supervised finetuning}
      \label{tab:afrimgsmzeroshot}
    \end{table*}
}

\newcommand{\insertSampledData}{

\begin{table*}[!t]
     \begin{center}
     \small
     \renewcommand{\arraystretch}{1.4}
     \resizebox{\textwidth}{!}{
    \begin{tabular}{ll}
    \toprule
    \textbf{Persona}  & \textbf{Prompt}  \\
    \midrule
    \multirow{4}{*}{\makecell[l]{A middle-aged male chef from Kumasi, intrigued by \\ Mahama's background in hospitality management, and \\ seeking to combine culinary skills with business acumen \\ to open a chain of restaurants.}}  & \multirow{4}{*}{\makecell[l]{Mahama dey go buy tomatoes wey e cost \$15, yam wey \\ e cost \$20, and fish wey e cost \$25 for market. \\ How much Mahama dey spend for market? \\ \emph{\color[HTML]{14AA34} \textbf{Language: Nigerian Pidgin (pcm)}}}}  \\
    &  \\
    &  \\
    &  \\
    \midrule
    \multirow{4}{*}{\makecell[l]{A young Yoruba filmmaker from Ibadan, passionate about\\ promoting indigenous languages through television series \\ and documentaries, inspired by Tunde Oladimeji's work.}}  & \multirow{4}{*}{\makecell[l]{Lehin ti o ra atupa kan ti o ni owo \$45 lati ile itaja kan, \\ Akin ti gba awon note meji ti \$20 ati nkan kan ti \$5 bi yiyan re \\ Elo ni Akin ni tele? \\ \emph{\color[HTML]{14AA34}{Language: Yoruba (yor)}}}}  \\
    &  \\
    &  \\
    &  \\
    \bottomrule
    \end{tabular}
    }
    \caption{Table showing selected personas, and generated prompts in different languages from the collection.}
    \label{tab:dataset_examples}
    \end{center}
    \end{table*} 
}

\usepackage[hang,flushmargin]{footmisc}

\usepackage{float}

\restylefloat{table}

\usepackage[T1]{fontenc}
\usepackage[utf8]{inputenc}

\usepackage{parskip}
\newcommand{\cmark}{\ding{51}}%
\newcommand{\xmark}{\ding{55}}%

\newcommand\Hau{\texttt{hau}\xspace}
\newcommand\Ibo{\texttt{ibo}\xspace}
\newcommand\Swa{\texttt{swa}\xspace}
\newcommand\Yor{\texttt{yor}\xspace}
\newcommand\Zul{\texttt{zul}\xspace}
\newcommand\Afripersonahub{\texttt{AfriPersonaHub}\xspace}
\newcommand\Afripersonainst{\texttt{AfriPersona-Instruct}\xspace}
\title{Improving Multilingual Math Reasoning for African Languages}

\author{%
Odunayo Ogundepo $^{1,2*}$, Akintunde Oladipo $^{1,2*}$, Kelechi Ogueji$^{1,2*}$,\\ \textbf{Esther Adenuga $^{1*}$, David Ifeoluwa Adelani $^{3,4*}$, Jimmy Lin $^{2}$.}\\
\footnotesize
$^1$The African Research Collective,
$^*$Masakhane NLP, 
$^2$University of Waterloo, \\
\footnotesize
$^3$Mila, McGill University, $^4$Canada CIFAR AI Chair, \\
\footnotesize
\texttt{Correspondence:\{ogundepoodunayo\}@gmail.com}}

\begin{document}
\maketitle
\begin{abstract}

Researchers working on low-resource languages face persistent challenges due to limited data availability and restricted access to computational resources. Although most large language models (LLMs) are predominantly trained in high-resource languages, adapting them to low-resource contexts, particularly African languages, requires specialized techniques.
Several strategies have emerged for adapting models to low-resource languages in today’s LLM landscape, defined by multi-stage pre-training and post-training paradigms. However, the most effective approaches remain uncertain. This work systematically investigates which adaptation strategies yield the best performance when extending existing LLMs to African languages. We conduct extensive experiments and ablation studies to evaluate different combinations of data types (translated versus synthetically generated), training stages (pre-training versus post-training), and other model adaptation configurations. Our experiments focuses on mathematical reasoning tasks, using the Llama 3.1 model family as our base model.

\end{abstract}

\section{Introduction}

Large Language Models (LLMs) owe much of their success to vast amounts of carefully curated, high-quality training data \cite{kaplan2020scaling,touvron2023llama,penedo2024the}.
However, assembling such data, particularly for underrepresented languages, remains a significant bottleneck due to data scarcity, privacy concerns, and the high cost of human annotation \cite{nekoto-etal-2020-participatory,singh-etal-2024-aya}.
To address these limitations, researchers have turned to a variety of alternative data creation strategies, including translating existing datasets from high-resource languages and prompting LLMs to generate synthetic data \cite{liu2024best,ge2024scaling,long-etal-2024-llms}.
Each of these approaches carries its own tradeoffs.
Translation can introduce literalism and semantic drift, particularly in specialised domains like mathematical reasoning, while also reinforcing source-language biases \cite{chironova2014literalism,al2022challenges,singh-etal-2024-aya}.
Synthetic data generation, in contrast, enables tailoring to low-resource contexts but raises questions about factuality, reasoning correctness, cultural fidelity \cite{ok-etal-2025-synthetic}, and linguistic accuracy of the generated text.
Both synthetic data generation and translation have been applied to improve instruction-following capabilities of LLMs \cite{ustun-etal-2024-aya,uemura-etal-2024-afriinstruct}, but their comparative effectiveness is still relatively unknown, especially for low-resource languages.

In this work, we systematically examine the effectiveness and trade-offs of these strategies for mathematical reasoning in African languages—a particularly challenging task for LLMs, given that their pretraining data is often skewed toward high-resource languages and may contain little to no mathematical content in African languages \cite{adelani2025irokobenchnewbenchmarkafrican}. We focus on supervised fine-tuning, which is a crucial part of the modern LLM post-training pipeline \cite{ouyang2022training,kumar2025llm,tie2025survey}, and has been shown to be effective for mathematical reasoning tasks \cite{yu2023metamath,luo2023wizardmath}. We use a multi-stage framework that includes persona-driven prompt generation, LLM-based synthesis, translated adaptations of existing math datasets, and monolingual vs multilingual training. Through a series of targeted experiments and ablations, we quantify how each approach impacts model performance across dimensions such as reasoning accuracy, resource efficiency, generalization, linguistic coherence, and cross-lingual transfer.

Some key findings and contributions of our work are as follows:
\begin{itemize}
    \item We apply a practical persona-based pipeline to generate synthetic data for mathematical reasoning in 9 low-resource African languages.
    \item We perform extensive comparative analysis between synthetically generated data and machine translated data, revealing their limitations and demonstrating their effectiveness across multiple languages seen and unseen during training. Our models approach or exceed frontier systems like GPT-4 in some languages.
    \item Contrary to common practice, we empirically show that masking prompts during SFT loss computation yields worse performance on mathematical reasoning for low-resource languages, suggesting that the extra loss signal could help the model understand the languages better.
    \item We perform several other ablations such as examining the effects of scaling, comparing monolingual vs joint multilingual training, and exploring the effects of continual pretraining, uncovering novel insights that could benefit builders of low-resource language technology.
\end{itemize}

We also release two key artifacts of our work: a.) \Afripersonahub: A collection of $45,000$ personas curated from Wikipedia and other web data sources. and b.) \Afripersonainst: IFT dataset containing $60,000$ elementary math problems in $9$ African languages generated through persona-driven data synthesis.
Overall, our work offers practical insights into the development of modern LLMs for low-resource languages and highlights promising directions for scalable data construction.

\insertSampledData

\section{Related Work}
Large Language Models (LLMs) have become increasingly ubiquitous in our daily lives.
These models are data hungry, especially with their growing scale and adaptation \cite{kaplan2020scaling}.
Hence, despite their growing popularity, they are seldom developed for low-resourced languages, especially African ones \cite{ahia-etal-2021-low-resource,joshi-etal-2020-state}.
To fill the data shortage gap, approaches such as translation, direct synthetic data generation and sample-efficient methods have been explored\cite{wang2025multilinguallanguagemodelpretraining, doshi-etal-2024-pretraining}.
Synthetic data, which refers to data generated by machine learning models
or algorithms, as opposed to directly by humans, has proven to be a viable alternative in recent times.
%Several methods have been introduced for generating synthetic data across various pipelines of LLM development.
Techniques have been developed for pretraining \cite{Gunasekar,tinystories}, post-training \cite{alpaca,orca} and even evaluation \cite{wei2024long}.
Synthetic data has also been applied to various domains such as multimodality \cite{sun2023aligning}, and math \cite{luo2023wizardmath,yu2023metamath}.
%Ror a more comprehensive overview of synthetic data generation methods, we refer the reader to various survey papers \cite{long-etal-2024-llms,liu2024best}.
Despite the rising popularity of these methods, there has been limited focus on their suitability for low-resource languages.
Other works have focused on translating existing datasets \cite{muennighoff2022crosslingual,kramchaninova-defauw-2022-synthetic,lai-etal-2023-okapi,aryabumi2024aya,ustun-etal-2024-aya,aakanksha-etal-2024-multilingual}.
However, translation could induce artifacts that adversely affect performance \cite{koppel2011translationese,dutta-chowdhury-etal-2022-towards,lai-etal-2023-okapi}.
Another approach has been to directly use large language models to generate responses in specific target languages \cite{aryabumi2024aya,ustun-etal-2024-aya,odumakinde2024multilingualarbitrageoptimizingdata,dang-etal-2024-rlhf}, but this not been explored for African languages.
Sample efficient methods that aim to squeeze out optimal performance from existing data have also been explored to significant success \cite{ogueji-etal-2021-small,dossou2022afrolm,adebara2022serengeti,ogundepo2022afriteva,oladipo2023better}.
Our work is orthogonal to these as we focus on answering the question of what the best approach is to train a \textit{modern} LLM that works on African languages.
Focusing on supervised fine-tuning for mathematical reasoning, we comprehensively compare data generation methods such as translation and synthetic data generation, as well as explore dimensions such as scaling and cross-lingual transfer. Our work offers novel insights for practitioners looking to build for these languages.

\section{Methodology}
\label{methodology}

This section describes our methodology for generating multilingual mathematical reasoning data in African languages.

\subsection{Synthetic Data Pipeline}
\label{synthetic-data}

Our aim is to generate synthetic data in a specified target language from a Large Language Model (LLM). 
A key attribute of good training data for LLMs is diversity.
Achieving a highly diverse dataset is no mean feat, even with human annotators \cite{liupopdistribution,parrish-etal-2024-diversity}.  
It is particularly challenging with synthetic data as ensuring diversity in LLM generation is not trivial \cite{bukharin-etal-2024-data,chen2024diversity,long-etal-2024-llms,liu2024best,gandhi-etal-2024-better,chung-etal-2023-increasing}.
One method that has bolstered diversity in synthetic data generation is the use of personas in the data synthesis prompt \cite{ge2024scaling} and it is proven to be effective in practice \citep{lambert2025tulu3pushingfrontiers}.

We utilize the text-to-persona and persona-to-persona pipelines introduced in \cite{ge2024scaling}.
We provide a piece of text, such as a Wikipedia page or a news article, and task GPT-4o to generate the persona of a person who might be interested in the given text, as well as their spoken language and country.
In another stage, we ask the model to generate a set of personas that could interact with a provided persona.
We use GPT-4o\footnote{GPT-4o-2024-08-06} to generate the personas, and supply text articles from Wikipedia and WURA \cite{oladipo2023better}. Note that the persona is asked to be generated in English language and we only use the first 200 words from the article page.
The specific prompt details can be found in the prompt in Appendices \ref{prompt:personageneration} and \ref{prompt:persona2persona}. We deduplicate our initial collection of persona using MinHash and Locality Sensitive Hashing (LSH). We use a shingle size of 3 and a similarity threshold of 0.8 to determine if two personas are duplicates.

We specify the target task as math, as well as add some extra constraints.
For example, we ask that the generated prompt should require between 2 and 8 steps to solve, and require the use of only basic arithmetic operations.
The specific prompt details can be found in the prompt in Appendix \ref{prompt:mathpromptgeneration}.
In the data synthesis prompt, we ask the model to utilize the provided persona to generate a math problem. We ask that the problem be generated in the specified target language. The data synthesis prompt can be found in Appendix \ref{prompt:mathproblem}.
We ask GPT-4o to provide a step-by-step solution to the given math problem in the language of the problem. The specific prompt used can be found in \ref{prompt:mathresponse}.
Data is generated for Nine African languages - Yoruba, Igbo, Hausa, Swahili, isiZulu, Nigerian Pidgin, Somali, Afrikaans and Arabic. These languages span Western, Eastern, Northern and Southern Africa, and have around a billion active speakers \footnote{https://www.ethnologue.com/browse/names/}.
\autoref{tab:dataset_examples} shows examples of generated personas and sample prompts generated for each persona.

\subsection{Translated Data Pipeline}
\label{translation-data}
We translate BigMath \cite{albalak2025bigmathlargescalehighqualitymath}\footnote{https://huggingface.co/datasets/SynthLabsAI/Big-Math-RL-Verified}, a high-quality verifiable mathematics dataset, and OpenMathInstruct V2 \cite{toshniwal2024openmath}\footnote{https://huggingface.co/datasets/nvidia/OpenMathInstruct-2}, an instruction-tuning mathematics dataset generated using Mixtral \cite{jiang2024mixtral}.
All translations are performed using GPT-4o with the translation prompt detailed in \ref{prompt:translationprompt}.
For both datasets, we focus on the grade school mathematics subset, randomly sampling prompt--response pairs and translating them into one of our target languages.
To ensure diversity across languages, we implement a constraint that prevents any prompt from being sampled more than once across all target languages.

\insertafrimgsmtable

\begin{figure*}[t]
    \centering
    \includegraphics[width=\linewidth, height=12cm]{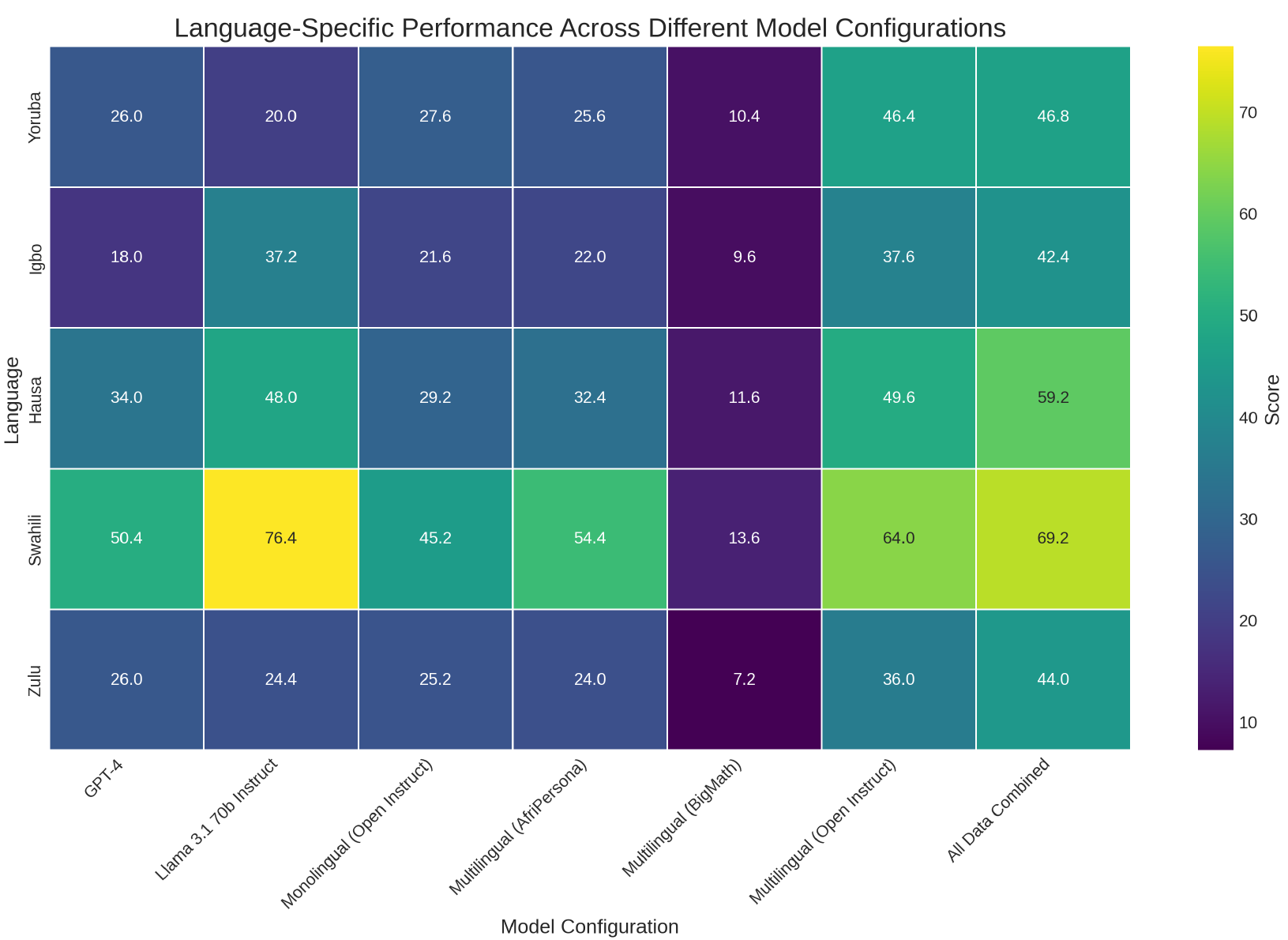}
    \vspace{-3mm}
    \caption{Heatmap showing performance across different model configurations for each language. This heatmap highlights the performance across Llama 3.1 8b instruct finetuned on different datasets}
    \label{fig:variability}
\end{figure*}

\section{Experiments}

\subsection{ Training Setup }
\label{sec:training}

Our methodology employs controlled experiments where we vary model and data configurations and evaluate their effectiveness.
All experiments utilize the Llama 3.1 model family \cite{grattafiori2024llama}, specifically the 8B parameter variants (\texttt{base} and \texttt{instruct}) due to computational constraints. For select ablation studies, we use the Lugha-Llama 8B model \cite{buzaaba2025lugha}, a Llama 3.1 base model continually pretrained on unsupervised African language data.
Our dataset comprises grade school level mathematical reasoning examples from synthetically generated examples using the pipeline described in \autoref{synthetic-data}, which we designate as \Afripersonainst and translations of BigMath and OpenMathInstruct described in \autoref{translation-data}.

Each dataset comprises an equal number of examples in $9$ languages (Afrikaans, Arabic, Hausa, Igbo, Nigerian Pidgin, Somali, Swahili, Yoruba \& Zulu).
We focus exclusively on supervised fine-tuning and fine-tune all models for 2 epochs with an effective batch size of 128 distributed across four(4) A100 80GB GPUS using DeepSpeed ZeRO-3 training optimization.
Training was performed in mixed precision using a learning rate of $5e^{-5}$, following a linear learning rate schedule that includes a warm-up phase spanning 3\% of the total training steps.
Following \citet{lambert2025tulu3pushingfrontiers}, we sum the loss across training tokens rather than averaging, which proved more effective for our objectives.

\insertafrimgsmtablezeroshot

\subsection{Evaluation}

We evaluate all models on AfriMGSM~\cite{adelani2025irokobenchnewbenchmarkafrican}, a multilingual test collection of $250$ parallel questions translated into $16$ African languages.
AfriMGSM is a human-translated version of GSM8K \cite{cobbe2021gsm8k}, which consists of grade school math word problems.
Although AfriMGSM only covers $5$ of the languages in our training dataset: \Hau, \Ibo, \Swa, \Yor, \Zul), we report the zero-shot effectiveness of our models on the remaining $10$ African languages in AfriMGSM.

The traditional evaluation method for assessing mathematical capabilities in language models relies on Exact Match Accuracy, which checks whether the correct answer appears within the model's generation.
However, we discovered that this approach proves inadequate when evaluating responses in multilingual contexts due to variations in response styles across languages and models.
Instead, we implemented an LLM-as-judge evaluation framework \cite{stephan2025calculationadjudicationexaminingllm}, which provides a more robust assessment method.
This approach prompts a larger and more capable language model to determine whether a generated response is correct with respect to the ground truth answer.
Specifically, we utilize GPT-4o as our evaluation judge, implementing this through the \texttt{lighteval} toolkit \cite{lighteval}.

To prevent data contamination between the training dataset and the evaluation benchmark, and also to reduce inflated benchmark scores \cite{li-etal-2024-open-source}, we de-duplicate our training data and AfriMGSM test samples using Locality Sensitive Hashing (LSH). 

\subsection{Key Ablations}

Our primary research objective is to determine the optimal combination of base model, training data, and training methodology for effectively adapting contemporary large language models to diverse African languages, with mathematical reasoning serving as our domain of investigation.
Given the linguistic diversity and low-resource nature of African languages, understanding these optimization factors becomes crucial for developing robust multilingual mathematical reasoning capabilities.
Through our comprehensive experimental framework, we systematically investigate several key research questions outlined below.

\subsubsection{Translation vs Direct Synthetic Data Generation}

Here, we try to answer the question: "Does translating high-quality English mathematical instruction datasets yield better model performance compared to generating synthetic data directly in target African languages?"
To answer this question, we conduct a comparative analysis between models trained on translated versions of established datasets (BigMath and OpenMathInstruct) and models trained on \Afripersonainst, our dataset generated directly in target languages.
To control for model quality as a confounding variable, we employ the same foundation model for both translation tasks and synthetic data generation, ensuring that performance differences are attributable to the data generation methodology rather than variations in the underlying generation model

\subsubsection{Prompt and Instruction Masking}

The impact of computing loss over prompt tokens has been explored in recent research on supervised fine-tuning, with findings suggesting that computing loss on the instruction portion of examples can be beneficial when instruction-tuning data is limited \cite{shi2024instructiontuninglossinstructions} or when instructions are significantly longer than the outputs \cite{huerta2024instruction}.

Inspired by these findings, we examine how different loss masking strategies impact model performance in low-resource settings. Specifically, we evaluate the effect of masking instruction and prompt tokens during training loss computation. Our investigation aims to determine whether restricting loss calculation exclusively to the model's outputs improves learning efficiency and generalization capabilities in multilingual, instruction-tuned models—particularly for low-resource African languages that constitute only a minimal portion of the overall training data.

\subsubsection{Scaling Data Size}

We also explore the effect varying the size of the training dataset across three configurations: 10,000, 20,000, and 30,000 examples.
The main aim is to see if we get increased downstream performance as a scale up the data size.
For each configuration, we maintain consistent training hyperparameters (as detailed in Section~\ref{sec:training}) and fine-tune the LLaMA 3.1 Instruct model. This allows us to isolate and quantify the specific effect of training data volume on model performance across the target African languages.

\subsubsection{Monolingual Experts versus Multilingual Generalists}

We investigate whether dedicated language-specific models yield superior performance compared to a single multilingual model trained on multiple African languages simultaneously. For this experiment, we utilize the translated Open Instruct dataset—which has demonstrated optimal downstream performance in our previous tests—and partition it by language to create separate training sets. Our approach involves fine-tuning the LLaMA 3.1 8B Instruct model under two distinct paradigms: (i) monolingual fine-tuning, where we train individual models exclusively on data from a single target language, and (ii) multilingual fine-tuning, where we train one comprehensive model on data from all five African languages concurrently. 

\subsubsection{Continual Pre-training}

Using an existing model, we explore the effect of continual pretraining on a model's mathematical reasoning capabilities in African languages. Continual pre-training represents an effective approach for adapting language models to new languages or domains \cite{muller-etal-2021-unseen,alabi-etal-2022-adapting,yildiz2024investigating,shi2024continual} and has consistently demonstrated benefits for mathematical performance \cite{shao2024deepseekmathpushinglimitsmathematical}. For this ablation study, we employ Lugha-LLaMA (8B) \cite{buzaaba2025lugha}, a model specifically adapted to African languages through pretraining on 6B tokens from the WURA corpus \citep{oladipo-etal-2023-better} spanning 16 African languages, complemented by 4B tokens from the English-language OpenWeb Math Corpus \citep{paster2024openwebmath}. Through this experiment, we investigate how such specialized continual pre-training influences models' ability to acquire mathematical reasoning skills in African languages through supervised fine-tuning.

\section{Results and Discussion}

\subsection{Qualitative Analysis of AfriPersona-Instruct}

We conducted a manual audit on a sample of 100 prompts from \Afripersonainst. This audit assessed the prompts' readability, clarity, and linguistic accuracy, guided by established grammatical rules. We found several issues that affected the naturalness and comprehension of the prompts. One such problem is unnatural phrasing where expressions did not align with typical Yoruba usage. Lexical inaccuracies were also common, such as the use of non-standard words unfamiliar to native speakers and words used in inappropriate contexts, thereby altering the intended meaning of the phrases or sentences. These lexical issues, coupled with grammatical errors, led to unnatural phrasing. 

Some prompts were ambiguous due to various factors, including unrelated adjacent sentences, missing leading questions that would provide context, incorrect word order, and various lexical issues. These problems contributed to the lack of clarity in the prompts' meaning.

These linguistic shortcomings likely affected the quality of the responses generated. Although the persona-based approach contributes to diversity in the generated samples, the identified issues highlight the need for refinement to ensure alignment with linguistic norms and natural usage of these languages.

\subsection{Effect of Prompt Masking}

To assess the impact of prompt masking, we compare rows 8 and 9 in \autoref{tab:afrigmsmresult}, which differ only in whether prompt masking is applied. Both settings use the same model (Llama 3.1 8b Instruct), dataset (\Afripersonainst), and number of synthetic examples (30,000), providing a controlled comparison.

We observe that masking prompts during loss computation (\texttt{Row 9}) leads to slightly worse performance than computing loss over the full input (\texttt{Row 8}), with the average score dropping from 39.4 to 38.7 overall, and from 31.8 to 29.7 when excluding English and French. These results suggest that, in our multilingual and low-resource setup, retaining the instruction tokens in the loss function is slightly more effective. Preserving the instruction in the loss signal may help models better ground their responses and generalise in these languages. Thus, our results refine prior conclusions, highlighting that masking may not always benefit multilingual, low-resource instruction tuning.
Based on these observations, we chose to keep the computed loss over the prompt tokens for subsequent experiments.

\subsection{Translated vs Direct Synthetic Generation}

We compare models trained on machine-translated versions of BigMath and OpenMathInstruct (\texttt{Rows 10–11}) in \autoref{tab:afrigmsmresult} against those trained on \Afripersonainst (\texttt{Rows 8–9}), under equivalent training budgets of 30k samples.

% \red{This paragraph still needs work.}
We find that translated data underperforms compared to direct generation when the source dataset is narrow in scope or specialized, such as BigMath. As shown in \texttt{Row 11}, training on translated BigMath yields a very low average score (11.2), and an even lower average on African languages (8.96), indicating limited generalizability of mathematical domain data across languages and formats.

In contrast, using the translated OpenMathInstruct dataset (\texttt{Row 12}) substantially improves performance, achieving a high overall average (51.4) and a strong average for African languages (46.8). This suggests that the breadth and diversity of the source dataset play a critical role in how well translated data performs.

Models trained on \Afripersonainst (\texttt{Rows 6–9}) show significant improvements over the vanilla LLaMA 3.1 8b Instruct model (\texttt{Row 3}). Training on 30k \Afripersonainst samples (\texttt{Row 8}) achieves a much higher 39.4 overall average and 31.8 average on African languages—more than 2× gain over the vanilla model, which achieves only 22.5 average across all languages and 14.6 on African languages.
% This demonstrates the strong utility of directly generated synthetic data for improving multilingual math reasoning in low-resource languages.

While these scores still fall behind models trained on high-quality translated data like OpenMathInstruct (\texttt{Row 11}, 51.4 overall), it is competitive with GPT-4's performance (\texttt{Row 1}, 40.4 overall and 30.9 on African languages). This highlights the effectiveness of native-language synthetic data, particularly in closing the gap to frontier models using relatively modest resources.

Training on all available data (i.e., both direct and translated high-quality sources) yields the best overall results, achieving the highest average (57.7) and strongest performance on African languages (52.3). This suggests that the two strategies are complementary: high-quality translated data can bootstrap multilingual capabilities, while directly generated native-language data strengthens performance and cultural alignment.

% Conclusion: While translation is a viable and scalable strategy, particularly when using broad instruction datasets like Open Instruct, our findings highlight the unique value of directly generating synthetic data in target African languages. The best results are achieved by combining both approaches, reinforcing the need for investment in native-language data creation alongside translation.

\subsection{Effect of Scaling Data Size}

We analyse the impact of increasing synthetic data size on downstream performance by fine-tuning the LLaMA 3.1 8B Instruct model with 10k, 20k, and 30k samples of synthetic data drawn from \Afripersonainst. As shown in \autoref{tab:afrigmsmresult}, performance steadily improves with scale across almost all African languages. Moving from 10k to 20k samples yields modest gains across \Ibo (17.6 → 22.0), \Hau (28.0 → 32.0), \Swa (40.8 → 43.6) and \Zul (15.6 → 20.8). A further increase to 30k samples produces more substantial gains, particularly for Yorùbá (25.6), \Swa (54.4), and \Zul (24.0), with the overall average on African languages improving from 24.2 to 31.8.

Compared to GPT-4 (\texttt{Row 1}), the 30k-trained model (\texttt{Row 8}) approaches parity on Yorùbá and Hausa (25.6 and 32.4 vs GPT-4’s 26.0 and 34.0, respectively).
On \Ibo and \Swa, the 30k-trained model outperforms GPT-4 (22.0 and 54.4 vs 18.0 and 26.0, respectively). Notably, the LLaMA 3.1 8B Instruct baseline (\texttt{Row 3}) lags significantly behind across all languages, with an average of only 22.5 compared to the 34.9 achieved by the 10k-trained model. Thus, even modest amounts of instruction-tuned synthetic data can substantially improve multilingual mathematical reasoning in low-resource African languages.

\subsection{Effect of Continual Pre-training}

We evaluate the impact of continual pre-training on downstream mathematical reasoning by comparing LLaMA 3.1 8B models with and without language adaptation prior to supervised fine-tuning. Specifically, we compare standard LLaMA 3.1 8B Base and the Lugha-LLaMA 8B model, which has undergone continual pre-training on African language corpora \cite{buzaaba2025lugha}, both trained on the same translated OpenMathInstruct dataset comprising 30k examples.

Lugha-LLaMA (\texttt{Row 15}) averages 38.8 overall and 36.0 without English/French, trailing the baseline’s 48.9 and 44.6 (\texttt{Row 14}). This suggests that while continual pretraining on African languages boosts general cross-lingual performance \citep{buzaaba2025lugha}, it may not directly benefit multilingual math reasoning under limited supervision. A likely reason is the distributional shift from literary/news pretraining data to math-focused fine-tuning. We hypothesize that domain mismatch limits performance, and suggest exploring domain-aligned pretraining to better support task-specific learning.

\subsection{Monolingual Experts versus Multilingual Generalists}

Comparing the results of finetuning separate models per language (\texttt{Row 5}) and finetuning a single multilingual model for all languages (\texttt{Row 11}).
The multilingual model outperforms the monolingual models by a large margin for all languages.
This suggests that joint multilingual training benefits from cross-lingual transfer, which is absent in monolingual training.

% \red{This still needs some work. Our story is: Mono generally does better if multilingual sees <20k samples}

% Due to the way our dataset is generated—with a fixed number of samples per language in the multilingual setup—monolingual models are trained on approximately 3,000 samples each. In contrast, the multilingual models see up to 30,000 samples in total. Despite this, the monolingual Yoruba model achieves a strong score of 27.6, outperforming the multilingual model trained on 10k samples (19.2) and even matching the multilingual 20k model (19.2) in Yoruba. However, the multilingual model significantly outperforms the monolingual models in higher-resource languages like Swahili and Zulu and shows consistent gains as the training set scales.

% These findings suggest that while monolingual fine-tuning may be competitive in lower-resource scenarios or for less frequently represented languages, multilingual training provides better average performance across all languages, and is especially beneficial for underrepresented ones. This makes multilingual fine-tuning a more scalable and generalizable approach.

\subsection{Generalisation to Unseen Languages}

To evaluate the generalisation, we test performance on a held-out set of African languages not seen during fine-tuning. \autoref{tab:afrimgsmzeroshot} presents accuracy scores on 10 such languages for various models and fine-tuning settings.

We observe that zero-shot performance improves significantly with task-specific supervised fine-tuning. The vanilla Llama 3.1 8B Instruct model performs marginally better than the base model, with gains in languages such as Kinyarwanda (Kin) and Luganda (Lug). However, once fine-tuned with 30,000 synthetic samples from the \Afripersonainst dataset, performance improves across almost all unseen languages, especially in Xhosa (Xho) and Shona (Sna), indicating effective transfer of mathematical reasoning capabilities.

The strongest generalisation is achieved by fine-tuning on the translated OpenMathInstruct dataset, with the Llama 3.1 8B Instruct model achieving the highest zero-shot accuracy on 8 out of 10 unseen languages. Notably, the model reaches 28.0\% accuracy on Xhosa, and consistently improves in low-resource languages like Afaan Oromo and Twi, showcasing the utility of synthetic multilingual instruction data.

These results indicate that fine-tuning on well-translated, diverse instructional data improves in-language performance (\autoref{tab:afrigmsmresult}) and substantially boosts generalisation to typologically and geographically diverse African languages (\autoref{tab:afrimgsmzeroshot}. This suggests that cross-lingual transfer in instruction-tuned models is feasible and beneficial for low-resource mathematical reasoning tasks.

% \subsection{Alignment Tax}

% \red{Odunayo, can you eval the best model we have on MMLU? We want to talk about any regression on Instruct model performance that we get from adapting for reasoning in African languages}

\section{Conclusion}

Focusing on mathematical reasoning, we have comprehensively explored the viability of various synthetic data generation pipelines for low-resource languages. Our findings demonstrate the effectiveness of synthetic instruction-tuned data for improving mathematical reasoning in low-resource African languages. Additionally, we perform several ablations across dimensions like prompt masking, monolingual vs multilingual training, and continual pretraining.
Overall, our results highlight a clear path forward: combining targeted multilingual data generation with strategic fine-tuning enables practical, high-performance instruction-following in low-resource languages.

%Directly generated native-language data (\Afripersonainst) significantly improves performance over baseline models and even approaches or exceeds frontier systems like GPT-4 in some languages. While high-quality translated datasets such as OpenMathInstruct yield strong results, especially in zero-shot generalisation, they are best leveraged in combination with direct synthetic generation to balance linguistic fidelity with task coverage.

%Contrary to common practice, prompt masking slightly reduces performance in multilingual low-resource settings, suggesting that preserving the full prompt during training better grounds model responses. Scaling the amount of synthetic data yields steady gains, and even modest instruction tuning (e.g., 10k samples) substantially boosts model performance over untuned baselines. However, continual pre-training on general African corpora does not consistently improve downstream mathematical reasoning, underscoring the need for domain-aligned adaptation.

%Finally, multilingual models trained on diverse data outperform monolingual experts in overall coverage and scalability, though monolingual fine-tuning remains competitive for specific languages. Overall, our results highlight a clear path forward: combining targeted multilingual data generation with strategic fine-tuning enables practical, high-performance instruction-following in low-resource languages.

\section{Limitations}

While our study demonstrates encouraging progress in enhancing mathematical reasoning in African languages through instruction fine-tuning, several limitations remain.

First, we do not formally verify the synthetically generated reasoning data used for training. Although the data is produced using high-quality prompting strategies, there is no guarantee of correctness or consistency across examples. Future work should include automated or human-in-the-loop verification to ensure data quality.

Second, our work focuses exclusively on mathematics. It is unclear whether the patterns we observe, such as the effects of scaling data size or instruction tuning across languages, would generalise to other domains like commonsense reasoning, reading comprehension, or scientific QA. Expanding to additional domains would help validate the robustness of our findings.

Lastly, our experiments rely on a single synthetic data generation setup. Exploring alternative prompt templates, source tasks, or instruction styles could provide insights into the stability and transferability of the results under different conditions.

\section{Acknowledgements}
This research was supported in part by the Natural Sciences and Engineering Research Council (NSERC) of Canada. 
Additional funding is provided by Microsoft via the Accelerating Foundation Models
Research program. 

\bibliography{custom}

\clearpage
\onecolumn
\appendix

\section{Challenges in Evaluating Mathematical Reasoning}

Evaluating the mathematical capabilities of generative language models in multilingual contexts presents unique challenges, stemming from the complexity of their free-form outputs and the difficulty of reliably comparing these outputs to ground truth answers.
The challenge of answer extraction is particularly pronounced when models embed numerical answers within paragraphs of explanation, use varied notational conventions, or express the same mathematical concept in multiple equivalent forms (e.g., "1/2" versus "0.5" versus "50\%").
We notice that these difficulties are even more pronounced in multilingual contexts.

One strategy often used when evaluating multiple choice benchmarks is to look at the LLM-estimated probability of the ground truth answer as a proxy for measuring the model capabilities.
However, log-likelihood evaluation has some limitations and does not correlate strongly with model capabilities \citep{fu2025multiplechoicequestionsreasoning}. 

To work around these issues, we use LLM-as-judge to measure the accuracy of a given response to the ground-truth answer.
Here we provide the judge with a question, the correct answer and the model generation. 
The judge is prompted to give a boolean response of 0 or 1 with a justification for the answer.
The prompt used in this evaluation is provided in Appendix~\ref{prompt:llmjudge}.
To ensure high accuracy, we calibrate the prompt using a couple of samples from different languages.

\section{Prompts}
\subsection{Persona Generation Prompt}
\label{prompt:personageneration}

\begin{lstlisting}[
  basicstyle=\ttfamily,
  breaklines=true,
  breakatwhitespace=true,
  breakindent=0\dimen0,
  columns=flexible,
  keepspaces=true,
]
Generate diverse user personas from an African context based on a given piece of text, such as a Wikipedia page or a news article, focusing on individuals who might be interested in the content.

- Analyze the key topics and content within the text with an emphasis on cultural, social, and economic contexts of Africa.
- Identify potential interests, needs, or challenges that a variety of users from diverse African backgrounds might have in relation to the text.
- Consider comprehensive demographic and psychographic characteristics such as age, profession, interests, ethnicity, cultural contexts, etc.

# Steps

1. **Content Analysis**: Break down the text to identify primary topics, themes, and connections to African contexts.
2. **Persona Generation**: For each theme, create diverse personas by:
   - Specifying a broad range of demographic characteristics, avoiding repetitive structures
   - Identifying relevant professions or industries prevalent in Africa
   - Outlining specific interests or hobbies related to the text and befitting diverse African cultures
   - Highlighting potential challenges or motivations relating to the content and African context
3. **Persona Detailing**: Write a brief description for each persona, including how they would engage with the content and their potential impact or interaction within the African context.

# Output Format

Express each user persona in a structured JSON format with the following attributes:
```json
{
  "countries": ["Kenya"],
  "languages": ["Swahili"],
  "persona": "A passionate advocate for digital innovation in Kenya, exploring solutions in mobile finance, deeply rooted in cultural traditions."
}
```
Include languages and countries as lists, and ensure the persona is detailed and contextually relevant.

# Examples

**Example Input:** Short excerpt from a text about renewable energy developments.

**Example Output:**
```json
[
  {
    "countries": ["South Africa"],
    "languages": ["English"],
    "persona": "An environmental enthusiast from South Africa, fervent about eco-friendly innovations and promoting sustainable practices in local communities."
  }
]
```

(Real examples should include 2-3 personas per text with varied demographic details, emphasizing diverse African contexts.)

# Notes

- Consider both direct and indirect interests, such as professionals in related fields or laypeople with hobbies connected to the topic.
- Aim for a broad spectrum, covering different age ranges, cultural backgrounds, socio-economic statuses, and levels of familiarity with the topic, specifically within African societies.
- Respond with the personas in English text, reflecting the diverse linguistic landscape of Africa
\end{lstlisting}

\subsection{Persona-to-Persona Generation Prompt}
\label{prompt:persona2persona}

\begin{lstlisting}[
  basicstyle=\ttfamily,
  breaklines=true,
  breakatwhitespace=true,
  breakindent=0\dimen0,
  columns=flexible,
  keepspaces=true,
]
Generate a list of 3 personas for each input persona in JSON only, considering interpersonal relationships and cultural diversity within Africa.

Identify new personas that have interpersonal relationships with existing personas, and generate additional personas that are similar but from different communities or countries in Africa.

## Steps

1. **Analyze the Given Persona**: 
   - Identify key characteristics such as country, language, role, and interests.
   - Understand potential interpersonal relationships based on shared activities or values.

2. **Generate Interpersonal Personas**:
   - Create personas that could interact or share goals with the given persona. Consider roles that would naturally support or collaborate with their interests or goals.

3. **Create Similar Personas from Different African Communities**:
   - Maintain the essence of the original persona but adapt cultural, geographic, or linguistic attributes to fit another African context.

4. **Diversity of Contexts**: 
   - Ensure a variety of regions, languages, and cultural perspectives are represented to reflect diverse African contexts.

# Output Format

- Provide a list of 3 personas in JSON format, each with fields: "countries", "languages", and "persona".
- Ensure that each persona includes a narrative description in the "persona" field that captures the relationships or adapted cultural details.

# Examples

## Given Persona:
```json
{
  "countries": ["Kenya"],
  "languages": ["Swahili"],
  "persona": "A passionate advocate for digital innovation in Kenya, exploring solutions in mobile finance, deeply rooted in cultural traditions."
}
```
### Generated Personas (3 examples):
```json
[
  {
    "countries": ["Kenya"],
    "languages": ["Swahili"],
    "persona": "An experienced mobile app developer in Kenya, collaborating with digital innovators to create culturally relevant financial solutions."
  },
  {
    "countries": ["South Africa"],
    "languages": ["Zulu"],
    "persona": "A forward-thinking entrepreneur in South Africa, bridging the tech divide with localized content creation and digital financial tools."
  },
  {
    "countries": ["Nigeria"],
    "languages": ["Yoruba"],
    "persona": "An enthusiastic promoter of technological adaptation in Nigeria, researching community-based mobile applications, grounded in traditional customs."
  }
]
```

# Notes

- When creating interpersonal personas, consider roles such as collaborators, supporters, or community leaders.
- When generating similar personas in different countries, adapt the technological focus and community values to local contexts.
- Aim for realism and cultural specificity in each persona.
- Generate only one JSON object
\end{lstlisting}

\subsection{Prompt Generation}
\label{prompt:mathpromptgeneration}

\begin{lstlisting}[
  basicstyle=\ttfamily,
  breaklines=true,
  breakatwhitespace=true,
  breakindent=0\dimen0,
  columns=flexible,
  keepspaces=true,
]
Create a prompt or an instruction to perform a task that a given user persona that is a native speaker of a given language might ask you to do relating to a particular topic of interest in a particular style

An example of an instruction in {seed_language} could be:
{seed_prompt}

Note:
1. The above example is not tied to any particular persona, but you should create one that is unique and specific to the given persona.
2. The instruction should contain all the verifiable contraint(s)
3. Your output should start with "User instruction:" 
4. Your output should not include the answer to the instruction
5. Your output should be in the provided language
6. Provide the prompt JSON format, each with fields: "prompt", "language"
\end{lstlisting}

\subsection{Math Problem Generation}
\label{prompt:mathproblem}

\begin{lstlisting}[
  basicstyle=\ttfamily,
  breaklines=true,
  breakatwhitespace=true,
  breakindent=0\dimen0,
  columns=flexible,
  keepspaces=true,
  mathescape=true
]
Create a math problem that a given user persona, who is a native speaker of a given language, might ask you to solve. Ensure the problem requires between 2 and 8 steps to solve, and the solution involves performing a sequence of elementary calculations using basic arithmetic operations ($+$, $-$, $\times$, $\div$) to reach the final answer.

An example of a math problem in {seed_language} could be:

{seed_prompt}

Note:
1. You should make full use of the persona description to create the math problem to ensure that the math problem is unique and specific to the persona.
2. Each math problem created should require between 2 and 8 steps to solve, involving elementary calculations using only basic arithmetic operations.
3. Your response should not include a solution to the created math problem.
4. Provide the prompt JSON format, each with fields: "prompt", "language"
5. Your output should be in the provided language
\end{lstlisting}

\subsection{Math Response Generation}
\label{prompt:mathresponse}

\begin{lstlisting}[
  basicstyle=\ttfamily,
  breaklines=true,
  breakatwhitespace=true,
  breakindent=0\dimen0,
  columns=flexible,
  keepspaces=true,
  mathescape=true
]
Provide a step-by-step solution to the given math problem in the language of the problem and write the final answer in a new line.

Note:
Ensure that all steps of your solution are written in the provided language and conclude by writing the final answer clearly on a new line.
\end{lstlisting}

\subsection{LLM-as-a-Judge Prompt}
\label{prompt:llmjudge}

\begin{lstlisting}[
  basicstyle=\ttfamily,
  breaklines=true,
  breakatwhitespace=true,
  breakindent=0\dimen0,
  columns=flexible,
  keepspaces=true,
  mathescape=true
]
You will be provided with a mathematics question, its golden answer, and the model's answer. The question and the model's answer could be in any language. Your task is to judge if the model's answer is equal to the provided golden answer,


Your task is to judge if the model's answer is correct or not based on the golden answer.
You should first briefly give your reasoning process regarding how the model's answer matches the golden answer(s), and then give the boolean score of 0 or 1 if they are equal or not.

----------------------------------------------------

Example 1:

Question: Jason had 20 lollipops. He gave Denny some lollipops. Now Jason has 12 lollipops. How many lollipops did Jason give to Denny?

Golden Answer: 8

Model's Answer: Jason started with 20 lollipops, but now he only has 12, so he gave Denny 20 - 12 = 8 lollipops.
Your Judgment: The model and the groundtruth answer are the same. The score: [[1]]
\end{lstlisting}

\subsection{Translation Prompt}
\label{prompt:translationprompt}

\begin{lstlisting}[
  basicstyle=\ttfamily,
  breaklines=true,
  breakatwhitespace=true,
  breakindent=0\dimen0,
  columns=flexible,
  keepspaces=true,
  mathescape=true
]
You are provided with a math problem and the step by step answer in english to that problem, your task is to translate the math problem and the provided response into a given language.

Note:
- Provide the prompt JSON format, each with fields: "problem_translation", "step_by_step_response"
- Your output should be in the provided language.
- Ensure that you Preserve the entity names that are mentioned in the original math problem during translation
- Ensure that you preserve numbers, mathematical formulas or symbols that are provided in the original prompt in the response
- Ensure that you only return the translations without any additional information or explanations

Math Problem: {{prompt}}
Answer: {{answer}}
Language: {{language}}
\end{lstlisting}

\end{document}